\documentclass[authorversion]{acmart}

\AtBeginDocument{%
  \providecommand\BibTeX{{%
    \normalfont B\kern-0.5em{\scshape i\kern-0.25em b}\kern-0.8em\TeX}}}

\setcopyright{acmlicensed}
\copyrightyear{2024}
\acmYear{2024}
\acmDOI{XXXXXXX.XXXXXXX}

\acmJournal{TOMM}
\acmVolume{1}
\acmNumber{1}
\acmArticle{1}
\acmMonth{3}

\usepackage{algpseudocode}%
\makeatletter
\def\BState{\State\hskip-\ALG@thistlm}
\makeatother
\usepackage{textcomp}
\usepackage{array}

\usepackage{bm}
\usepackage{multirow}
\usepackage{epsfig}
\usepackage{subfig}
\usepackage{subcaption}
\usepackage{graphicx}
\usepackage{pifont}
\usepackage{xcolor}
\usepackage{tabularx}
\usepackage{colortbl}
\usepackage{svg}

\newcommand{\cmark}{\ding{51}}%
\newcommand{\xmark}{\ding{55}}%

\newcommand{\etal}{\textit{et al.}}

\newcommand{\eg}{\textit{e.g.}}

\begin{document}

\title{Leveraging Speech for Gesture Detection in Multimodal Communication}

\author{Esam Ghaleb}
\orcid{0000-0002-0603-9817}
\affiliation{%
  \institution{University of Amsterdam}
  \streetaddress{Science Park 904}
  \city{Amsterdam}
  \state{North Holland}
  \country{The Netherlands}
  \postcode{1098 XH}
}
\email{e.ghaleb@uva.nl}

\author{Ilya Burenko}
\affiliation{%
  \institution{ScaDS.AI Dresden/Leipzig}
  \country{Germany}
}
\affiliation{%
  \institution{TU Dresden}
  \country{Germany}
}

\author{Marlou Rasenberg}
\affiliation{%
  \institution{Meertens Institute}
  \country{The Netherlands}
}
\author{Wim Pouw}
\author{Ivan Toni}
\affiliation{%
  \institution{Radboud University}
  \country{The Netherlands}
}

\author{Peter Uhrig}
\affiliation{%
  \institution{ Friedrich-Alexander-Universität Erlangen-Nürnberg}
  \country{Germany}
}
\author{Anna Wilson}
\affiliation{%
  \institution{University of Oxford}
  \country{United Kingdom}
}

\author{Judith Holler}
\author{Asl\i~Özy\"{u}rek}
\affiliation{%
  \institution{Radboud University}
  \country{The Netherlands}
}
\affiliation{%
  \institution{Max Planck Institute for Psycholinguistics}
  \country{The Netherlands}
}

\author{Raquel Fern\'{a}ndez}
\affiliation{%
  \institution{University of Amsterdam}
  \country{The Netherlands}
}
\email{raquel.fernandez@uva.nl}

\renewcommand{\shortauthors}{Ghaleb et al.}

\begin{abstract}
Gestures are inherent to human interaction and often complement speech in face-to-face communication, forming a multimodal communication system. An important task in gesture analysis is detecting a gesture's beginning and end. Research on automatic gesture detection has primarily focused on visual and kinematic information to detect a limited set of isolated or silent gestures with low variability, neglecting the integration of speech and vision signals to detect gestures that co-occur with speech. This work addresses this gap by focusing on co-speech gesture detection, emphasising the synchrony between speech and co-speech hand gestures. We address three main challenges: the variability of gesture forms, the temporal misalignment between gesture and speech onsets, and differences in sampling rate between modalities. Our approach leverages a sliding window technique to handle variability in gestures' form and duration, using Mel-Spectrograms for acoustic speech signals and spatiotemporal graphs for visual skeletal data. We investigate extended speech time windows and employ separate backbone models for each modality to address the temporal misalignment and sampling rate differences. We utilize Transformer encoders in cross-modal and early fusion techniques to effectively align and integrate speech and skeletal sequences. The study results show that combining visual and speech information significantly enhances gesture detection performance. Our findings indicate that expanding the speech buffer beyond visual time segments improves performance and that multimodal integration using cross-modal and early fusion techniques outperforms baseline methods using unimodal and late fusion methods. Additionally, we find a correlation between the models' gesture prediction confidence and low-level speech frequency features potentially associated with gestures. Overall, the study provides a better understanding and detection methods for co-speech gestures, facilitating the analysis of multimodal communication.
\end{abstract}





\maketitle

\section{Introduction}
\label{sec:intro}
Gestures are a fundamental component of human interaction, serving many functions such as illustrating objects or actions, emphasising and further delineating verbal communication, as well as conveying deictic expressions \cite{mcneill1992hand}. 
Gesture detection, a key aspect of a wider project of automatic gesture analysis, aims to identify the onset of a hand gesture.
Current gesture detection approaches in human-computer interaction and human behavior analysis have two main limitations.
First, they primarily focus on a \textit{finite set of gestures with limited variability.} These gestures usually represent actions, objects, or conventionalised gestures (e.g., the thumbs-up sign) and are produced silently, without concurrent speech \cite{benitez2021ipn, molchanov2016online}.
However, in face-to-face interaction, which is the main form of communication, gestures usually appear in conjunction with speech. 
Such co-speech gestures stand in a complex interplay with spoken language, carrying various communicative purposes beyond the stand-alone meanings of silent gestures \cite{kendon2004gesture, wagner2014gesture, holler2019multimodal, holler2020communicating, holler2011co}. Second, the prevalent approach to automatic gesture detection uses mainly kinematic information, e.g., collected through motion capture technology \cite{guo2021human}, or visual information obtained from RGB recordings \cite{molchanov2016online, zhu2018continuous, kopuklu2019real, benitez2021ipn, ghaleb2023co}. Thus, despite the evident relationship between speech and gestures, in most approaches \textit{the speech information is neglected}.

\begin{figure}
    \centering
    \includegraphics[width=0.99\linewidth]{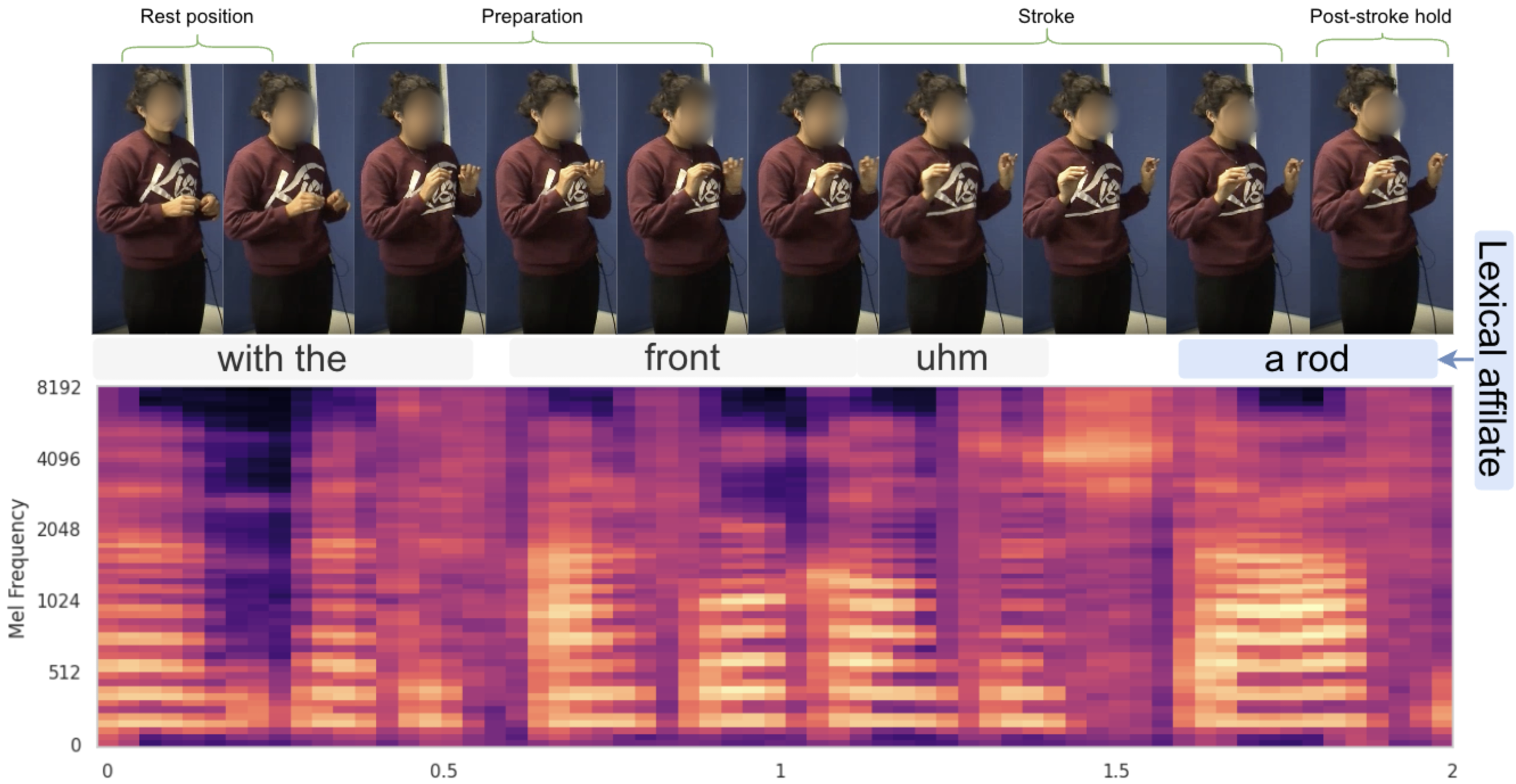}
    \caption{The figure illustrates the interaction between gestures and speech over two seconds. The speaker starts in a rest position and begins the gesture unit with a preparation phase. This is followed by the stroke phase, which is the meaningful part of the gesture unit. The gesture stroke is semantically related to the accompanying speech, i.e., ``rod'', sometimes referred to as the lexical affiliate. The speaker ends the gesture unit with a post-stroke hold. Typically, this is followed by a retraction phase, and then the speaker returns to the rest position again. We work with co-speech gestures that vary in form and duration based on the accompanying speech. }
    \label{fig:gesture_and_speech}
\end{figure}

In this work, we present an approach to gesture detection that addresses these limitations. We focus on manual co-speech gestures, which are semantically, pragmatically, and temporally
linked to speech. These gestures, along with speech, constitute an integral part of human language \cite{holler2011co}. Moreover, we emphasize gestures' synchrony with speech by proposing novel methods to integrate the speech and visual modalities. Figure \ref{fig:gesture_and_speech} shows an example of a co-speech gesture unit. The gesture unit consists of one or more movement phases, always including the gesture stroke (i.e., gesture onset), which is the most meaningful phase of the gesture \cite{kendon2004gesture}. The unit starts with a gesturing body part moving away from a resting position, optionally starting with a preparation phase, optionally followed by a pre-stroke hold, but always followed by a stroke. The stroke might then be succeeded by a post-stroke hold, a retraction phase, or both, indicating the completion of the gesture unit when the body returns to the resting position.
In this work, we focus on identifying strokes of iconic gestures, which are representational gestures and depict meaning. Figure \ref{fig:gesture_and_speech} shows an example of an iconic gesture unit that consists of preparation, stroke, and post-stroke hold. In this example, it is important to note a slight time difference between the beginning of the gesture stroke and the corresponding lexical affiliate, e.g., ``rod'' in the speech stream as shown in Figure \ref{fig:gesture_and_speech}. Therefore, even though speech and gestures are semantically and temporally related, incorporating the two modalities' data input into a unified framework presents \textit{several challenges}. 

\textit{First}, the form and duration of co-speech gestures vary depending on the speech they accompany, making their detection a difficult task. Our approach uses a sliding window with a short audio and video offset to enable consideration of different durations of gestures through sequential data-based analysis.
In the speech sequence, we use Mel-Spectrograms to represent the acoustic speech signals of each time window, which have been proven useful for various speech-related tasks such as speech and emotion recognition \cite{radford2023robust,ghaleb2021skeleton, hershey2017cnn}. 
In the vision sequence, we extract skeletal data from a visual time window and construct a spatiotemporal graph, which represents the dynamics of upper body and hand movements that have been shown to be effective in several tasks such as sign language and gesture detection \cite{yan2018spatial, song2020stronger, ghaleb2023co, ghaleb2021skeleton}.

\textit{Second}, the stroke onset of representational gestures often precedes the onset of their lexical affiliate by approximately 200-500 ms \cite{donnellan2022timing, ter2024hand}, reflecting a potentially systematic temporal off-set that characterizes the coordination between gestural and speech \emph{semantic} cues. Studies have shown that this coordination varies depending on speakers' familiarity with accompanying lexical affiliates and gesture types \cite{holler2019multimodal}.
Another well-known phenomenon in gesture-speech coordination takes place at the level of prosody, specifically showing that gestures with a beat-like quality tend to occur with prosodically marked speech. It is often found that beat-possessing gestures tend to co-occur more often and in a time-synchronised way with prosodically salient moments in speech, such as pitch accents \cite{wagner2014gesture}. 
It has been shown that there are biomechanical links of gesture to the respiratory-vocal system, which have been suggested to be a driver for gesture to co-occur with prosodically emphasized speech \cite{pouw2020energy}. 
Since representational gestures can also have a beat-like component, it is possible that gesture detection is informed by acoustic features that are related to prosody. Both the lexical affiliate and the multimodal prosody research predict that gesture is not randomly timed with speech, and there is some non-random relation between semantic and prosodic features of speech. Given that there is possibly an immediate and more long-range interaction of gesture with speech, we investigate the impact of using longer speech time windows to account for speech delay and 
coordinate the two modalities.

\textit{Third}, The frequency of communicative cues, such as speech and accompanying gestures, is different, with speech occurring more frequently than gestures. Moreover, the sampling rates for speech and visual input of co-speech gestures differ. For instance, speech is usually sampled at 44.1 kHz, while the frame numbers for co-speech gestures are different. These differences present a technical challenge. Hence, we employ speech and skeletal backbone models that handle each modality separately to produce embedding spaces with similar dimensions. However, as surveyed by Nyatsanga et al.~\cite{Nyatsanga2023}, speech and visual gesture embeddings show a limited correlation in the latent space. We employ Transformer 
encoders to tackle this challenge and further exploit sequential data obtained from the sliding window. These encoders process, contextualize, and align sequences of speech and skeletal embeddings, enabling the integration of these modalities through early and cross-modal fusion techniques. In summary, we make the following contributions:

\begin{itemize}
    \item In contrast to most current gesture detection approaches, our study focuses on identifying co-speech gestures, i.e., gestures that typically occur together with speech in communication. 
    \item We perform an analysis that reveals 
    significant differences in low-level frequency-based acoustic features when a gesture accompanies speech, which indicates that combining 
    speech and visual information is a promising direction for the gesture detection task.

    \item We implement and evaluate several multimodal fusion methods using Transformer encoders in order to improve the temporal alignment and contextualization of speech and skeletal sequences.
    
    \item Our study validates the approach on a naturalistic face-to-face interaction dataset. The results show that bimodal integration using cross-modal and early fusion outperforms several baseline methods, including unimodal and late fusion approaches. Specifically, our findings demonstrate that speech information enhances gesture detection, and expanding the speech buffer beyond visual time segments further improves the performance.
    \item We extensively analyse the study results. Our findings indicate that, when speech is employed, there is a correlation between models' confidence in gesture prediction and low-level speech features that are associated with speech accompanied by gesture.
\end{itemize}

The remaining sections of the paper are organized as follows. Section \ref{sec:relatedwork} provides a literature review on gestures, automatic gesture analysis with and without speech, and multimodal learning. Section \ref{sect:dataset} gives an overview of the dataset we use and details the processing steps to construct speech and vision sequences. In Section \ref{sect:speech_low_level_features}, we study whether speech information may include features predictive of co-occurring gestures. 
Sections \ref{sect:methods} and \ref{sec:exp_setup} describe how we employ early and cross-modal fusion methods with Transformer encoders to temporally align and contextualize speech and skeletal sequences, plus additional experimental details. 
In Section \ref{sect:results}, we present our results, which we further analyse in Section \ref{sect:analysis}. 
Finally, Sections \ref{sect:discussion} and \ref{sect:conclusion} discuss the implications of our findings and conclude the work.

\section{Related Work}
\label{sec:relatedwork}
\subsection{Gestures and Speech}
\label{sect:speech_and_gesture}
Co-speech gestures are meaningful physical movements of upper body parts, integral to language production and comprehension. 
In this overview, we briefly discuss co-speech gesture categories and their functions. 



\subsubsection{Gesture Categories}
Manual hand gestures can be classified into different categories, considering different criteria, e.g., their link to speech and whether they are voluntary or involuntary \cite{kendon2004gesture}. Here, we focus on a classification by McNeill (1992)  \cite{mcneill1992hand},  which categorises gestures into representational and non-representational gestures based on their relationship with speech \cite{mcneill1992hand, kendon2004gesture}. Firstly, representational gestures, including iconic and metaphoric gestures, are body movements deployed, e.g., to echo or elaborate the meaning of co-occurring speech. Iconic gestures visually represent speech's content by means of movements that resemble physical properties or actions of what is being said.
Metaphoric gestures extend this to abstract concepts, visually expressing metaphorical speech. 
The second category is non-representational gestures, which include deictic and beat gestures. Deictic gestures situate objects, locations, or persons within the communicative environment or abstract space. Beat gestures are short and repetitive rhythmic movements that do not carry semantic information. However, even iconic gestures may have a beat-like aspect to them, similar to beat gestures without representational content. Beat gestures and beat-possessing representational gestures, thus 
emphasize or stress certain parts of speech or signal important points in a discourse.

\subsubsection{Functions of Gestures}
Gestures contribute to the communication process in different forms, and they play a role in complementing or substituting speech in various ways \cite{wagner2014gesture}.
For example, they help regulate turn-taking during conversations. In ambiguous or complex speech, iconic gestures serve as a means of disambiguation, describing or emphasizing certain aspects of the spoken word. 
Emblematic gestures (i.e., gestures with conventionalised form and functionalities such as waving ``bye'') can even substitute speech entirely, conveying a message without the need for verbal expression. 
Through their rhythmic nature, beat gestures are particularly effective in emphasizing specific parts of speech, highlighting key points or indicating shifts in tone or topic. 
Some gestures have a complementary function. For example, saying ``I'd like some more'' while using a gesture to point at food on a table provides a visual cue that adds information to the verbal message. 
Furthermore, gestures can be more effective than speech for certain purposes, such as conveying objects' shapes. 
Gestures can facilitate speech comprehension in noisy settings \cite{drijvers2017visual}, and people enhance the gestural kinematics in noisy environments \cite{trujillo2021speakers}.

\subsection{Gesture Analysis}
\label{subsect:representations}
Gesture analysis is a key research area in human-computer interaction (HCI), sign language recognition (SLR), and behaviour analysis. It leverages various data types: \eg, sensory data gathered through technologies such as motion capture or glove-based sensors \cite{guo2021human}, visual data for gesture detection \& recognition \cite{kopuklu2019real,ghaleb2023co, zhu2018continuous}, and speech for gesture generation \cite{Nyatsanga2023}. This overview covers gesture detection and recognition studies/datasets and their use of speech information.

\subsubsection{Gesture Detection and Gesture Datasets}
In gesture detection and gesture recognition (i.e., classification), visual data-based 
models are currently the most dominant in the field \cite{kopuklu2019real, molchanov2016online, benitez2021ipn, zhu2018continuous}. 
In addition, the majority of datasets used for these tasks consist of people performing predefined gestures (e.g., depicting objects or performing actions) from a dictionary or silent gestures like a ``waving by'' gesture. An example of such a dataset is the ChaLearn Multimodal Gesture Recognition Challenge 2013 \cite{Escalera2013}. This well-established dataset includes various forms of data, such as audio, skeletal, and RGB with depth images. It contains $956$ video clips featuring $26$ participants, each performing $20$ standard gestures drawn from an Italian gesture dictionary.

The efficiency of detection and recognition models for these specific gesture categories has nearly reached its peak due to the task's simplicity. Acknowledging this, a recent study by Liu et al. \cite{Liu2022b} introduced a novel and more challenging dataset, namely LD-ConGR. It contains individuals performing specific gestures, however, in challenging settings. This dataset presents a promising direction for gesture recognition in HCI scenarios, where certain gestures could replace speech, especially in long-distance interactive scenes like virtual meetings or smart home environments. Despite the complexity of the LD-ConGR dataset, current state-of-the-art computer vision methods achieve a high recognition accuracy of 94\% \cite{Liu2022b},  indicating that the relative complexity of this data can largely be modelled by vision alone.
This suggests the need for a more challenging and general setup to advance the field. This highlights the relevance of our research that focuses on co-speech gestures, where their form and duration vary significantly depending on the accompanying speech and due to their idiosyncratic nature.

\subsubsection{Speech In Gesture Analysis}
The relationship between speech information and co-speech gestures has been investigated through gesture prediction and generation tasks. For example, Yunus et al. \cite{yunus2021sequence} used speech prosodic features such as F0 and intensity to predict gesture timing with a sequence-to-sequence model. They found that F0 is relevant to determining gesture timing.
A study by Pouw et al. \cite{pouw2020energy} found that hand gestures have biomechanical connections to the respiratory system, modulating speech and affecting its amplitude envelope. Kucherenko et al. \cite{kucherenko2022multimodal} found that gesture semantics and phase can be predicted better than chance using F0-related features and that this prediction can be generalised to new speakers. 

In gesture generation, as surveyed by \cite{Nyatsanga2023}, including information related to speech prosody and semantics improves the quality of generating beat and representational gestures, respectively, making them seem more natural according to raters' evaluations. An example pipeline of gesture generation is the work by Bhattacharya et al. \cite{bhattacharya2021speech2affectivegestures}, which employed a dual-model approach for the speech modality, utilising one model based on prosodic features and another that leverages word embeddings. By adopting such an approach, the study aimed to capture both speech's prosodic and semantic characteristics separately, which are closely related to co-speech gestures.

\subsubsection{Multimodal Gesture Analysis}
The notion of ``multimodality'' is frequently employed in many gesture detection and recognition studies to refer to 
learning approaches applied to different types of visual input. These visual inputs may consist of a sequence of RGB or depth images, skeletal data, or a combination thereof. Examples of this direction are evident in studies published between 2015 and 2019, as highlighted in \cite{wu2016deep, wang2017large, Abavisani2019}.
Following the release of the ChaLearn dataset in 2013 \cite{Escalera2013}, several studies initiated multimodal gesture recognition efforts, including inputs not only from vision but leveraging audio, video, and skeletal modalities. Each gesture in the dataset was accompanied by a word or a short phrase, pronounced by an actor while performing the gesture, that expressed the same meaning as the gesture itself. One known example is the work of Neverova et al. \cite{Neverova2015}, which introduced a multi-scale multimodal deep learning framework for gesture recognition and detection. 

In summary, while some progress has been made in multimodal approaches to gesture analysis, these efforts have largely been restricted to handling a dictionary of gestures. 
This study develops a multimodal framework for detecting co-speech gestures, addressing limitations in previous research. Specifically, we use a co-speech gesture dataset, which has been used recently for a vision-based gesture detection approach \cite{ghaleb2023co}. This dataset consists of face-to-face dialogues between two participants, requiring a more sophisticated analysis of gesture detection


\subsection{Multimodal Learning and Fusion}
Multimodal learning, which involves the fusion and integration of different data types like vision, text, and speech, is a popular research area in machine learning \cite{baltruvsaitis2018multimodal}. The most straightforward fusion of different modalities is ensemble averaging, e.g., where multiple models are created per modality, and their output is combined 
through averaging or weighted sum. However, this does not address a key challenge in this field: the alignment of modalities to ensure that information from different sources is effectively integrated. A well-known approach in the domain of aligning two modalities is based on contrastive learning, which aims to bring representations of different modalities closer and increase their similarity in the latent space. A representative example of this research direction is CLIP \cite{radford2021learning}.

Moreover, multimodal fusion and alignment can be achieved at a lower level, e.g., by integrating the embeddings at earlier layers. This became possible thanks to one important advancement in this area: the Transformer architecture, known for its flexibility and parallelizable nature \cite{vaswani2017attention}. Transformers excel at modelling long dependencies, which is crucial for tasks involving multiple modalities. This architecture has been successfully applied to various language and vision tasks, including visual question answering (VQA), cross-modal retrieval, and numerous multimodal recognition tasks. Specifically, Transformers are utilized in two main configurations for multimodal tasks: single-stream and multi-stream input systems \cite{khan2022transformers}. Single-stream input systems feed multimodal data into a single Transformer. For example, Visual-BERT \cite{li2019visualbert}, a variant in this category, employs a single Transformer on text and image embeddings to automatically model the relationship between the two modalities. It initially employs task-agnostic pre-training (such as predicting masked tokens from text input) before focusing on specific tasks (such as VQA). 
This single-stream setup is also used by Audio-Video HuBERT (AV-HuBERT) \cite{Shi2022a} for audio-video speech recognition, a task where the bimodal framework is used to transcribe speech from speech signals and lips. AV-HUBERT uses a ResNet to embed images of lips' movements and WAV2VEC \cite{baevski2020wav2vec} to embed speech. The embeddings are concatenated and fed to a Transformer encoder. The model is trained to predict masked elements of speech using the two input modalities.
In contrast, multi-stream Transformers, such as LXMERT
\cite{tan2019lxmert}, use separate Transformer Encoders for each modality. These include additional cross-modal encoders with cross-attention layers to model the interplay between visual and text streams.

Our work uses a single-stream system through a standard Transformed encoder and a multi-stream setup using a cross-modal encoder as proposed by \cite{tan2019lxmert}. We compare these two systems with a simple ensemble averaging approach, which develops separate speech and skeletal models and averages their output.

\section{Data and Preprocessing}
\label{sect:dataset}

This section gives an overview of the dataset used in our study, detailing the preprocessing steps to convert data from audio-video recordings into a sequential format. 


\subsection{Dataset}

We use the dataset developed by Rasenberg et al.~\cite{rasenberg2022primacy}, which contains 19 face-to-face, task-oriented dialogues involving 38 participants across 16 hours of recorded video. The dataset is different from other gesture datasets such as ChaLearn \cite{Escalera2013} or LD-ConGR \cite{Liu2022b}. While those datasets contain short video segments that show gestures linked to specific actions or items, the dataset by Rasenberg et al.~consists of face-to-face dialogues with co-speech gestures that occur naturally. These gestures can vary in form and duration depending on the accompanying speech. 
The dataset includes 6,106 manually annotated gesture strokes, with average and median durations of 0.58 and 0.42 seconds, respectively. Additional details on the annotated gesture strokes are provided by Ghaleb \etal~\cite{ghaleb2023co}, who employ this dataset for gesture detection.
Detailed studies on the nature of the collected data can be found in 
\cite{rasenberg2022primacy, eijk2022cabb}.

\subsection{Constructing Speech and Visual Sequences}
\label{sect:preprocessing}
Drawing on the approach employed by Ghaleb \etal~\cite{ghaleb2023co}, our study adopts a sliding window technique to create data sequences from participants' videos and audio recordings. We configure each window to include 15 frames, approximating the mean duration of a gesture stroke at a frame rate of 29.97 fps, which is 0.58 seconds. These windows are moved by an offset of 2 frames (67ms) each time. To illustrate this, a window starts at frame 0 and ends at frame 15; the next window starts at frame 2 and finishes at frame 17, continuing in this manner.
Subsequently, we transform these overlapping windows into distinct, non-overlapping sequences for our analysis. In our setup, a sequence is composed of 40 such windows. Taking the first sequence as an example, it starts with the window spanning frames 0 to 15 and includes every window up to the 40th, which, due to the 2-frame increment per shift, encompasses a total span of 96 frames. The sliding window approach results in a total of 21,220 speech and vision sequences and 848,800 time windows.

\subsubsection{Labeling} 
We adopt the labelling scheme from \cite{ghaleb2023co} by annotating a sliding time window as a $gesture$ stroke if it overlaps with the underlying segmented gesture stroke significantly (i.e., more than 50\%); otherwise, a time window is labelled as $neutral$. This results in 66,049 gesture and 782,751 neutral time windows, which account for $7.8\%$ and $92.2\%$ of the total samples, respectively. Due to the nature of the dataset, which contains natural conversations where substantial parts are without any gestures, it is expected that there will be an imbalanced distribution of labels. 
In Section \ref{sect:focal_loss}, we explain techniques to address class imbalance using specialised focal loss and subsampling during training.
The labels' set is denoted as follows: $C = \{Gesture (G), Neutral (N)\}$.


\subsubsection{Visual Input: Spatio-Temporal Graph}\label{sect:st_graph}
We extract speakers' pose (skeleton) from the RGB parts of the time-windows. Specifically, we use MMPose \cite{sengupta2020mm} to accurately detect and track body joints' 2D positions (i.e., the $x$ and $y$ coordinates) and a confidence value for the estimated positions. 
This results in $133$ body joints.
As proposed by Yan \etal~\cite {yan2018spatial}, we construct spatio-temporal graphs from the estimated poses. Each graph
consists of $j$ joints (i.e., vertices) and $e$ edges, $V = (j, e)$. A graph has two types of edges: spatial edges that comply with the natural connectivity of joints and temporal edges that connect the same joints across subsequent frames.

Hence, a time window of a skeleton (vision) sequence can be represented with a tensor as follows: $\bm{V} \in \mathbb{R}^{c \times t_v \times j}$, where $c$ is a joint data point (x, y, confidence), $t_v$ is the number of frames in a visual time window, and $j$ is the total number of body joints. We use $27$ of upper body joints, which have been shown to be the most relevant for sign language recognition \cite {jiang2021skeleton} and gesture detection \cite{ghaleb2021skeleton}. Finally, a vision sequence of $n$ time windows is denoted as: $\bm{V}^{(1:n)} \in \mathbb{R}^{n \times c \times t_v \times j}$.

\subsubsection{Speech Input: Mel Spectrograms}
We use Mel Spectrograms to represent speech signals of each time window. Spectrograms capture audio frequency features over short frames.
The Mel Frequency scale is used as it aligns with the perceived pitch. Moreover, Mel spectrograms have been used as a raw speech input for numerous speech-related tasks such as speech recognition (notably in Whisper \cite{radford2023robust}) and speech emotion recognition \cite{ghaleb2023joint}.

The Mel spectrograms are extracted as follows. First, we apply the Short-Time Fourier Transform (STFT) on the speech signal using a frame length of 25 ms, a stride of $10$ ms, and a periodic Hann window function to compute the spectrogram. We then convert the spectrogram frequencies into 64 Mel bands on a logarithmic scale. In our work, we use a sampling rate of 16 kHz for speech signals. Each speech signal in our setup has a duration of $500$ ms, resulting in $48$ overlapping frames per Mel spectrogram. Hence, a time window in a speech sequence can be represented with a matrix as follows: $S \in \mathbb{R}^{f \times t_s}$, where $f$ is the number of Mel bands, and $t_s$ is the number of frames in a spectrogram. 
As a result, a speech sequence of $n$ Mel spectrogram is denoted as a tensor as follows: $\bm{S}^{(1:n)} \in \mathbb{R}^{n \times f \times t_s}$. 
The Mel spectrogram sequences represent the speech input for our framework, providing a rich and efficient representation of speech characteristics that caputures variability in speech caused by glottal as well articulatory modulations.



\section{Preliminary Analysis: Is Speech Promising for Gesture Detection?}
\label{sect:speech_low_level_features}
We start by studying to what extent speech may include information that is predictive of co-occuring gestures. To this end, we conduct a preliminary analysis that investigates how low-level speech features are distributed when gestures accompany speech.
We restrict our attention to two types of low-level frequency representations.  First, using frequency analysis from the raw speech signal, we extract the fundamental frequency ($F0$).
$F0$ can be defined as the lowest dominant frequency of the speech waveform and is known as the ``pitch''. Sometimes, it is referred to as the first harmonic, where other harmonics are its multiples (e.g., $F2= 2 \cdot F0$). Second, from the frequency domain of the speech signal (speech spectral), we extract Mel-Frequency Cepstral Coefficients (MFCCs), which are the spectral representations of speech frequencies that allow for a representation of speech acoustics constrained by glottal and more upper-tract articulatory sources \cite{najnin2019speech}. 

\subsection{Data and Speech Frequency Features}
In this analysis, we sample $1$ second time windows that contain transcribed speech with at least one voiced segment and ensure equal representation across the two classes (gesture vs.~neutral). Additionally, we select speech samples that co-occur with gestures, controlling the voiced segments ranging from one to nine for each speaker. We then sample an equal number of speech samples without gestures (with an equal distribution of voiced segments). In total, we obtain 43,000 samples.

We extract the frequency features through the speech processing toolkit OpenSmile \cite{eyben2015opensmile}. This toolkit extracts $F0$ and MFCCs using a sliding window approach and offers a wide range of statistics. We use the following statistics: mean, maximum, skewness, and kurtosis. In addition, it offers a log of the ratio of $F0$ with respect to its first three harmonics (closely related to spectral tilt). When the ratio is high, it indicates a speech signal that is dominated by the fundamental frequency, characterized by a clearer or tonal sound (versus a less tonal, breathy, or creaky sound). Appendix \ref{app:speech_feat} includes a detailed list of the features.

\begin{figure}
    \centering
    \includegraphics[width=1\linewidth]{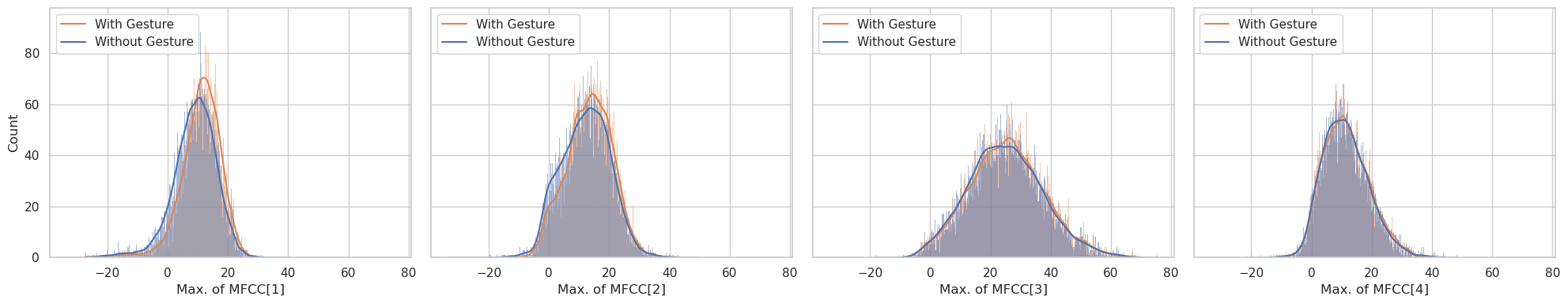}
    \caption{Distribution of Maximum of MFCCs features when speech occurs with or without gestures.
    The number of voiced segments is controlled to have a similar distribution with or without gestures.
    In the first figure, the distribution of the maximum of MFCC[1] is much higher when a gesture accompanies speech.}
    \label{fig:mfcc_max}
\end{figure}

\subsection{Analysis Results}
We observe that speech features, with few exceptions, vary significantly depending on whether speech is accompanied by gestures or not. Notably, for the maxima of the first and second MFCCs, we observe significant differences with extremely low $p$-values and highest t-test values ($t=31.1$ and $t=18.9$ for MFCC[1] and MFCC[2], respectively, $p \ll 10^{-5}$). The test values suggest a higher maximum MFCC[1] and [2] in the presence of gestures. These marked differences are shown in Figure \ref{fig:mfcc_max}.
The third and fourth coefficients remain statistically significant, with slightly higher t-test values ($t=5.4$ and $t=5.1$ for MFCC[1] and [2], respectively, with $p\ll 10^{-5}$). 
Regarding $F0$, we have two observations. First, the log of the ratio of $F_0$ in relation to its first three harmonics is significantly higher ($t=21.0$, $p\ll 10^{-5}$) when gestures accompany speech. Second, we note that apart from the $F0$ mean, other statistics of $F0$ are significantly different. Please refer to Appendix \ref{app:speech_feat} for a more detailed list of features, their statistics, and their variation in speech samples.

Based on this analysis, we conclude that speech is a promising source of information for gesture prediction. Concretely, we find that the spectral features of speech in the time and frequency domains vary significantly depending on whether speech co-occurs with gestures. In light of these findings, our gesture detection experiments employ a deep learning model that automatically learns frequency representations based on Mel spectrograms, which are more than hand-crafted features effective for speech tasks.
We explore how the confidence of gesture prediction through speech and bimodal models correlates with these observations (see Section \ref{sect:confidence_analysis}).
\section{Methodology and Models}
\label{sect:methods}
\begin{figure}
    \centering
    \includegraphics[width=0.99\linewidth]{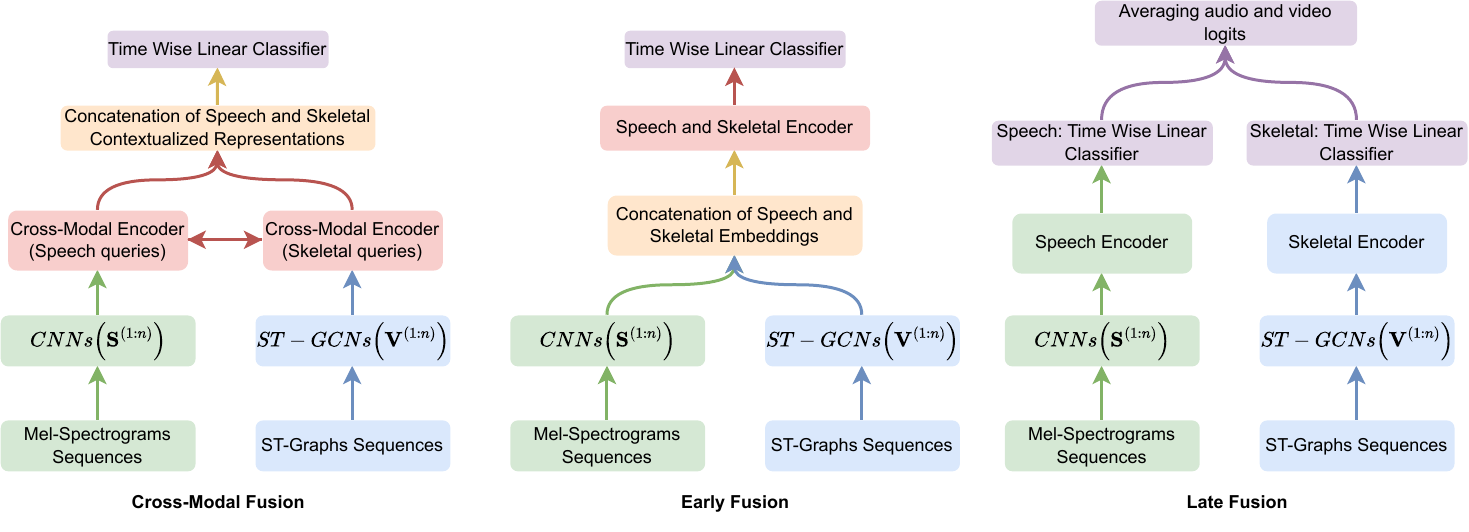}
    \caption{Diagrams of the employed models' architectures. Our approach progressively employs four steps. (1) It starts by preparing speech and vision sequences (explained Section \ref{sect:dataset}). (2) it then embeds two modal sequences using modality-specific models (see Section \ref{sect:embedding_models}). (3) The framework employs three fusion strategies—late, early, and cross-modal—to create contextualized unimodal, bimodal, or cross-modal embeddings, respectively. Late fusion combines separate modality predictions, while early and cross-modal fusions integrate both streams. (4) It applies a classification step on the embeddings per time window in the sequence.
    }
    \label{fig:framework}
\end{figure}

This paper presents a framework for detecting co-speech gestures from speech and visual information. The methodology we propose, which we describe in this section, uses two-stream network architectures as depicted in Figure \ref{fig:framework}. The process starts with preparing input speech and vision sequences, followed by embedding these sequences separately.
Finally, different fusion approaches are employed to encode and classify speech and vision sequences.

\subsection{Problem Definition} In our setup, each input sample can be denoted as a bimodal sequence pair: ($\{\bm{V}^{(1:n)}, \bm{S}^{(1:n)}\}, \bm{y}^{(1:n)}\}$), 
where $\bm{V}^{(i)}$ and ${\bm{S}}^{(i)}$ represent the ST-graph and Mel-spectrogram matrices at position $i$ in a visual and speech sequence, respectively,
and $y^{(i)}$ the corresponding label for that time window (i.e., gesture or neutral). As presented in Section \ref{sect:dataset}, the visual input, $\bm{V}^{(1:n)}$, is a sequence of a Spatio-Temporal (ST) Graph, while the speech input, $\bm{S}^{(1:n)}$, is a sequence of Mel-spectrograms. The objective of our task is to predict the joint conditional distribution, represented by $\bm{y}^{(1:n)}$, given the bimodal input sequence denoted by $(\{\bm{V}^{(1:n)}, \bm{S}^{(1:n)}\})$:
\begin{align}
    P(\bm{y}^{(1:n)}|\{\bm{V}^{(1:n)}, \bm{S}^{(1:n)}\}) = \prod_{i=1}^{n} P(\bm{y}^{(i)}|\{\bm{V}^{(1:n)}, \bm{S}^{(1:n)}\}).
\end{align}

We parameterize the joint conditional distribution using a Transformer-based Encoder that processes vision or speech sequences.
This means that the label prediction at a specific position $i$, denoted as $P(\bm{y}^{(i)})$, is based on the embeddings at that position. Nonetheless, these embeddings are derived from a contextualizing encoder that operates on the entire input sequence.

\subsection{Speech and Vision Models}\label{sect:embedding_models}

\subsubsection{Vision Model}
The vision model operates on the computed ST-graphs of time-windows sequences and uses Spatio-Temporal Graph Convolutional Networks (ST-GCNs) to represent the skeletal movements. ST-GCNs result in vision (skeletal) embeddings for each time window, represented by $\bm{v}^{(i)}$. For the entire sequence, the vision model embeds its time windows in the following manner: $\bm{v}^{(1:n)} = ST\text{-}GCNs(\bm{V}^{(1:n)})$. In our configuration, the visual embedding vector has 256 dimensions.

\subsubsection{Speech Model}
We employ a speech model that operates on Mel-frequency-based spectrograms.  Specifically, we utilize VGGish 
\cite{hershey2017cnn}, a Convolutional Neural Networks (CNNs) architecture, to obtain speech embeddings. VGGish, an adaptation of the VGG architecture, was also used for soundtrack classification \cite{abu2016youtube}.
We refer to speech embeddings of the corresponding time window as \textbf{$\bm{s}^{(i)}$}, where i stands for an $i^{th}$ time window in a speech sequence. For an entire sequence, this step embeds its Mel-spectrograms as follows: $\bm{s}^{(1:t)} = CNN(\bm{S}^{(1:t)})$. In our setup, the speech embedding vector has a similar dimension to the vision embeddings, namely $256$.

\subsection{Fusion Models}\label{sect:fusion}
We use three fusion methods: late, early, and cross-modal. They are applied at either the decision level (i.e., late fusion) or the embeddings level (early and cross-modal fusion). In this section, we provide a detailed explanation of these fusion techniques. We also describe the encoder architecture and how we use it to encode each modality separately in the late fusion method or to align and integrate both modalities in the early or cross-modal fusion methods.

\paragraph{Classifiers Architecture.} Our approach incorporates a classification network on the encoders' embeddings. This network contains two linear layers with a non-linear function in between, specifically the Rectified Linear Unit (ReLU). This network configuration of the classifier is consistently employed across the three fusion techniques.

\subsubsection{Late Fusion}
\label{sec:late}
The late fusion architecture consists of separate speech and vision encoders and two classifiers for each modality. The architecture is illustrated in Figure \ref{fig:framework}, right. 

\paragraph{Transformer Encoders.} We use the Transformer encoder \cite{vaswani2017attention} to obtain contextualised representations over entire sequences of speech or visual embeddings. In late fusion, one encoder is used for each modality separately. The encoder consists of a multi-head self-attention (MHSA) layer, followed by a position-wise feed-forward layer (FF). MHSA allows the model to focus on different parts of the input sequence, assigning more importance to certain time windows for detection or prediction tasks.
The MHSA layer improves the model's ability to learn rich representations at different positions from multiple subspaces. 
Sequences of time-window embeddings (i.e., vision or speech embeddings) are fed into the Encoder. For instance, for the visual embeddings, this step results in contextualised embeddings, which we denote as $\bm{u}_v^{(1:t)} = \textit{Encoder}(\bm{v}^{(1:t)})$. 

\paragraph{Classifiers.} The late fusion architecture has speech and vision classifiers. We ensemble the predictions of the two modalities' logits by taking their averages per time window.

\subsubsection{Early Fusion}
The early fusion approach integrates the speech and visual streams and aligns them over time before the decision level. It operates on the concatenated embeddings of the two modalities and uses a single Transformer Encoder. The model architecture is shown in Figure \ref{fig:framework}, center.
\paragraph{Transformer Encoder}
The concatenated embeddings are fed into an encoder with the same architecture as the late fusion encoders explained Section~\ref{sec:late}. The concatenation is applied on each time window's speech and vision embeddings. The encoder processes the embeddings, and the resulting contextualized embeddings are used for classification. This process can be represented as follows: $\bm{u}_{(s,v)}^{(1:t)} = \textit{Encoder}({concat(\bm{v}, \bm{s})}^{(1:t)})$.

\paragraph{Classifier.} The classifier utilizes the embeddings produced by the encoder.

\subsubsection{Cross-Modal Fusion}
Similarly to early fusion, cross-modal fusion is applied to the embeddings of the two modalities. However, in this case, fusion is achieved through a cross-modal encoder, which applies a cross-modal process to the embeddings of the two modalities.

\paragraph{Cross-modal Encoder.}
We use the cross-modal encoder proposed by Tan and Bansal \cite{tan2019lxmert}. It consists of a standard encoder (which has MHSA and FF) and is followed by three layers: Multi-Head Cross-Attention (MHCA), MHSA, and FF layers. The MHCA is a bidirectional attention layer (e.g., from speech to vision or vice versa). Unlike the MHSA layer, the MHCA layer employs the attention mechanisms cross-modally. In MHCA, a score is computed to determine how much emphasis to put on each part of the input sequence (e.g., speech sequence) given the other modality sequence (e.g., vision sequence). This is done using a function of the query (Q, i.e., from speech) and key (K, from vision) using a dot product: $\text{Score}(Q, K) = QK^T$.
This is followed by a score normalization using a softmax function. This step ensures that the attention weights are a valid probability distribution: $\text{Attention Weights} = \text{Softmax}(\text{Score})$.
The attention weights are then used to create a weighted sum of the values (V), another representation of the input data, which in our case are from the modality corresponding to the keys (i.e., from vision): $\text{Output} = \text{Attention Weights} \cdot V$.

Note that, in MHSA, queries, keys, and context vectors come from the same input sequence. In contrast, in MHCA, these vectors are derived from speech and vision embeddings. For instance, the queries come from the speech modality in a speech cross-modal encoder. Hence, speech embeddings are used to direct the attention mechanism towards relevant parts of the visual information. 

\paragraph{Classifier.} The resulting cross-modal embeddings per modality are concatenated and fed into the classifier. 


\subsection{Learning Objective and Handling Class Imbalance}\label{sect:focal_loss}
We optimise the framework by using focal loss (FL) as an objective function. Focal loss is an adaptation of cross-entropy (CE) loss ($CE(p^c) = -\log(p^c)$) and addresses class imbalance in tasks like ours, where most instances are neutral as explained in Section~\ref{sect:dataset}. It modifies the contribution of each sample to the loss based on the classification error using the formula:
\[
FL(p^c) = -\alpha^c (1 - p^c)^\gamma \log(p^c)
\] 
where \( p_c\) is the model's estimated probability for class \( c \). The parameters \( \alpha_c \) (scaling factor) and \( \gamma \) (focusing parameter) control the balance between classes and the rate at which easy examples are weighted during the training phase. With this loss, we ensure more focus on misclassified and difficult examples and mitigate the impact of a dominant class on model training.

Another step to address the class imbalance problem in our dataset is to employ a sub-sampling approach. Specifically, we use all sequences containing at least one gesture and balance them with an equal number of sequences consisting of neutral time windows. This process is applied in each epoch of the training phase.
The model is evaluated on the entire test dataset without sub-sampling to check how it generalizes to imbalanced data.

\subsection{Implementation Details}
We train our models on a computational node with 4 NVIDIA A100 GPUs. For the computationally most expensive cross-attention scenario, it takes approximately 1 hour for a model to converge on a single fold. We increase linearly the learning rate for the first 20 epochs until it reaches $10^{-4}$, and then we decrease the learning rate by a factor of 5 if our learning objective was not improving for 20 epochs. We find a batch size equal to 128 to be optimal for our experiments.

\begin{figure}
    \centering
    \includegraphics[scale=0.5]{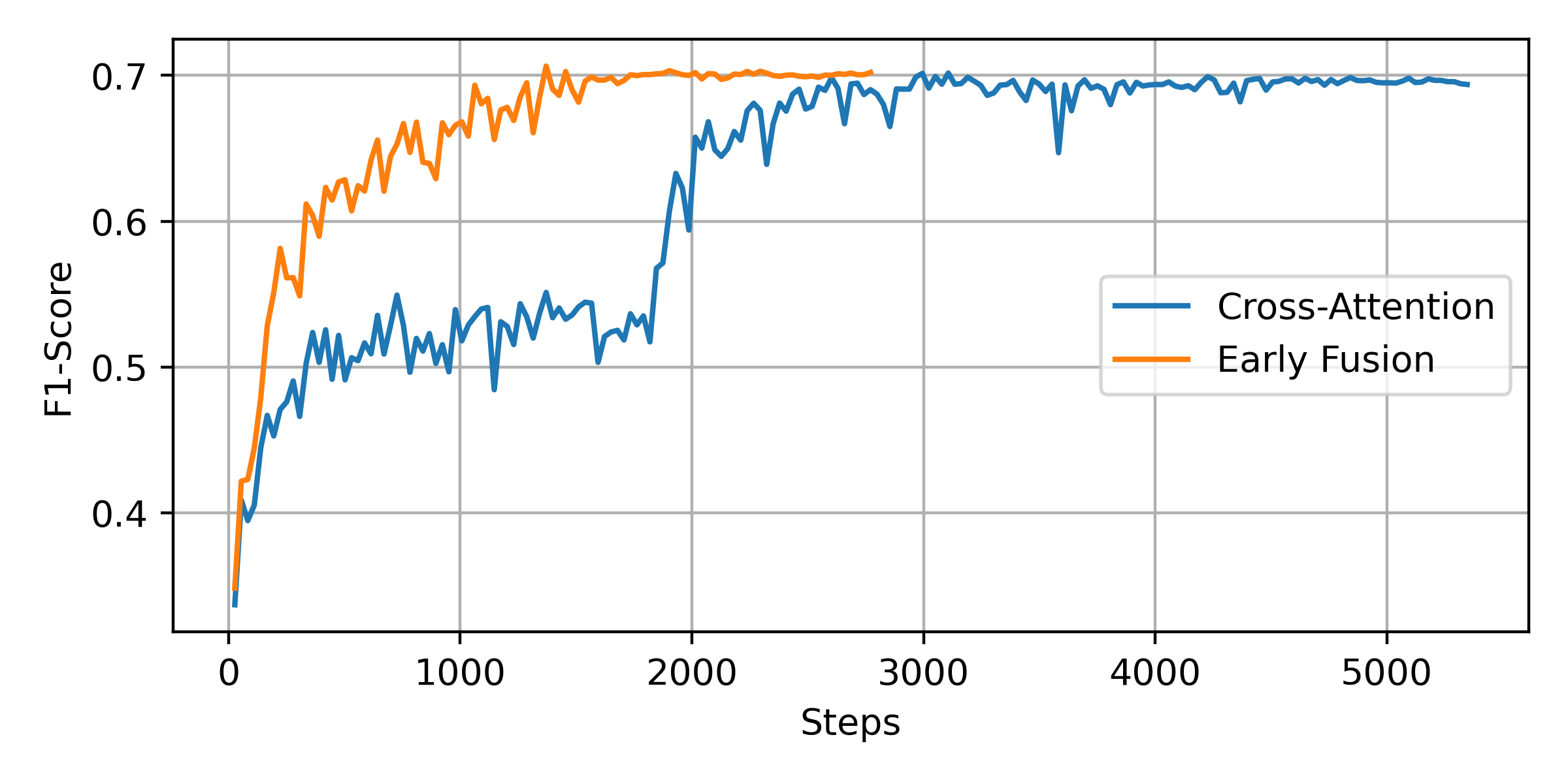}
    \caption{We trained models until convergence. Arguably, in the Cross-Attention scenario, F1-Score drastically increases in the middle of the training procedure after successfully combining inputs from both modalities.}
    \label{fig:ca_vs_ef_f1}
\end{figure}

We notice that the training curves in the early fusion and cross-attention scenarios look different. 
As shown in Figure \ref{fig:ca_vs_ef_f1}, in the latter case, we drastically improve classification quality in the middle of the training procedure. 
In the early fusion case, the self-attention mechanism receives as input concatenated features from pre-trained backbones, whereas in the cross-attention case, it takes much longer for the attention mechanism to adjust the product of matrices $Q$ and $K$ from one modality to the matrix $V$ obtained from the counterpart modality. The implementation is available in our GitHub repository.\footnote{Github project:  \url{https://github.com/EsamGhaleb/Bimodal-Co-Speech-Gesture-Detection}}

\section{Experimental Setup}
\label{sec:exp_setup}
We describe our evaluation setup, including metrics, baselines, and ablations. 
\subsection{Evaluation Metrics and Protocol}
We report the results using two key metrics: F1 score and Mean Average Precision (MAP). These metrics are explicitly used to evaluate the class imbalance challenge. The F1 score is the harmonic mean of precision and recall, while MAP is determined by making a binary decision across the full range of prediction probabilities.
These metrics are calculated per class, where we mainly focus on the class of interest: gesture. We also report the Receiver Operating Characteristic (ROC) and Precision-Recall (PR) curves for results analysis. These two metrics visualize the results of our models for the two classes over threshold values.

We use 5-fold cross-validation, partitioning the dataset into five distinct folds. In each fold, samples are allocated to the testing or training set, depending on the fold index.
The performance metrics reported in the results and analysis sections are the average scores computed over all five test folds. 

\subsection{Baselines and Ablations}
To demonstrate the performance of the proposed bimodal approaches in co-speech gesture detection, we compare them in two ways. First, we compare the fusion models against baseline unimodal models, i.e., models with either speech or vision alone. Additionally, we develop speech models with varying durations to account for the fact that gesture strokes typically precede their lexical affiliate by $200$ to $500$ milliseconds, as suggested by previous studies \cite{ter2024hand, donnellan2022timing}. To achieve this, we add extra speech buffers of $250ms$ and $500ms$ to the right of each window, extending the default speech time window from 500ms to 750ms and 1000ms. This results in Mel spectrograms with dimensions of $48 \times 64$, $72 \times 64$, and $96 \times 64$ for respective time windows. We train and evaluate speech models using data from the three buffers: 0, 250, and 500 milliseconds. 

Second, we check how different bimodal fusion models combine speech and visual input. Specifically, we investigate how late fusion, which is a simple ensembling approach, compares to early and cross-modal fusion, which include more advanced integration and alignment methods. This allows us to study the two key techniques we employ to align both modalities: heuristically using different speech buffers or by means of a Transformer encoder.

\paragraph{Sanity Check}
We assess whether the improvements observed with the bimodal fusion models come from the effective use of speech and vision or simply from the increased size of these models. To determine this, we substitute the speech embeddings with a random Gaussian noise vector while maintaining the original vision embeddings. Subsequently, all fusion models are trained and evaluated on the random vectors. We compare the resulting sanity-check models with the bimodal and vision-alone models to control for improvements that could be due to chance. 

Finally, we also report a majority-class baseline and a random baseline. The random baseline assigns a random prediction value between 0 and 1 for each time window, while the majority-class baseline predicts every sample as the most frequent class (i.e., the neutral class in our study). 
\section{Gesture Detection Results}
\label{sect:results}
In this section, we present the main results of the proposed procedures and compare them with unimodal baselines that include either vision or speech. 


\subsection{Vision and Speech Baselines}
\paragraph{Skeletal model.} 
The skeletal model achieves an F1 score of 66.2 and a MAP of 70.5, as shown in the first row of Table \ref{tab:all_vggish_results}. This model serves as the visual baseline for our subsequent results. This backbone model has already demonstrated its effectiveness in achieving high-performance metrics, as established by Ghaleb \etal~\cite{ghaleb2023co}. Building on this foundation, our current research is focused on examining the extent to which integrating speech signals into the framework can further augment the model's capability to detect gestures. First, however, we report the results obtained by the speech-only model.


\paragraph{Speech model.}
The speech model's results 
are shown in the row for the ``Speech alone'' approach in Table \ref{tab:all_vggish_results}. The results reveal two key insights. Firstly, detecting gestures from speech (using only speech acoustic information) is significantly better than the majority-class and random baselines (shown in the last two rows of Table \ref{tab:all_vggish_results}).
While, as expected, the speech-only model obtains lower results than the skeletal model, the performance of the speech models is clearly higher than that obtained by the baselines, with the F1 score varying between $38.9$ and $44.1$. Secondly, we observe that extending the speech buffer improved detection performance across the two metrics. For example, the F1 and MAP scores of the speech model improve from $39.0$ and $32.4$ to $44.1$ and $40.5$, respectively, when using a buffer of 500ms. 

These results demonstrate that gestures can be predicted from speech alone beyond mere chance levels. Moreover, the study highlights the significance of temporally aligning speech and gesture strokes, evidenced by the impact of a speech buffer beyond gesture windows, which significantly improves detection performance. Next, we show that speech signals complement visual signals for gesture detection, as evidenced by the gain in the detection performance when fusing both modalities.


\begin{table*}
\caption{Results of vision and speech model baselines and the fusion models with three speech buffers (500, 250, and 0.00 milliseconds). All the fusion models use visual signals (skeletons), while ``sanity check'' refers to the model variants where speech signals are replaced with Gaussian noise; ``--'' stands for not applicable.}
\begin{tabular}{llccc}
\toprule
\bf Approach & \bf Speech Enabled & \bf Buffer (ms) & \bf F1 & 
\bf MAP\\
\midrule \midrule
Vision alone (skeletal \cite{ghaleb2023co})& -- & -- & 66.2$\pm$1.2 & 70.5$\pm$1.2 \\
\midrule
\multirow{3}{*}{Speech alone} & \cmark & 500 & 44.1$\pm$0.5 & 40.5$\pm$1.6 \\
 & \cmark & 250 & 38.9$\pm$1.2 & 32.8$\pm$1.5 \\
 & \cmark & 0.00 & 39.1$\pm$1.2 & 32.4$\pm$1.0 \\
\midrule
\multirow{4}{*}{Cross-modal fusion} 
 &\cmark & 500& 69.5$\pm$1.0 & 
 73.1$\pm$0.7 \\
  &\cmark & 250 & 68.8$\pm$0.8 & 
 72.4$\pm$1.0 \\
  &\cmark & 0.00& 68.7$\pm$1.0 & 
 72.2$\pm$1.1 \\
  &\xmark~~(sanity check) & -- & 67.3$\pm$0.8 & 
 70.1$\pm$0.9 \\
\midrule
\multirow{4}{*}{Early fusion} 
  &\cmark & 500& \textbf{69.6$\pm$1.0} & 
 \textbf{74.2$\pm$1.1} \\
  &\cmark & 250 & 68.2$\pm$0.7 & 
 72.3$\pm$1.5 \\
  &\cmark & 0.00& 68.3$\pm$0.6 & 
 72.1$\pm$1.1 \\
  &\xmark~~(sanity check) & -- & 66.6$\pm$0.9 & 
 70.6$\pm$0.8 \\
\midrule
\multirow{4}{*}{Late fusion} 
  &\cmark & 500& 68.8$\pm$1.4 & 
 72.6$\pm$1.4 \\
  &\cmark & 250 & 67.0$\pm$1.4 & 
 71.3$\pm$0.9 \\
 &\cmark& 0.00& 67.1$\pm$0.8 & 
 71.1$\pm$1.3 \\
  &\xmark~~(sanity check) & -- & 66.4$\pm$0.6 & 
 70.5$\pm$0.9 \\
\midrule
Random baseline & -- & -- &13.5$\pm$0.5 & 7.8$\pm$0.3 \\
Majority class baseline & -- & -- & 0.0$\pm$0.0 & 7.8$\pm$0.3 \\
\bottomrule
\end{tabular}
\label{tab:all_vggish_results}
\end{table*}

\subsection{Speech and Vision Fusion}
Table \ref{tab:all_vggish_results} presents F1 and MAP scores for gesture stroke detection of various fusion methods and speech buffers. The results reveal several key points. \textit{First,} we see that combining speech and visual information through any fusion approach outperforms baseline methods that rely solely on speech or visual data. For instance, the cross-modal fusion approach with a speech buffer of 500 milliseconds achieves an F1 score of 69.5 and a MAP of 73.1. 
This is a statistically significant improvement compared to the vision-only skeletal model's F1 score of 66.2 and MAP of 70.5. These scores demonstrate the advantage of the bimodal approach in identifying gestures instead of depending only on visual data, as previously reported in Ghaleb \etal's study \cite{ghaleb2023co}.

\begin{figure}[t]
    \centering
    \begin{minipage}{0.3\linewidth}
        \centering
        \includegraphics[width=1.1\linewidth]{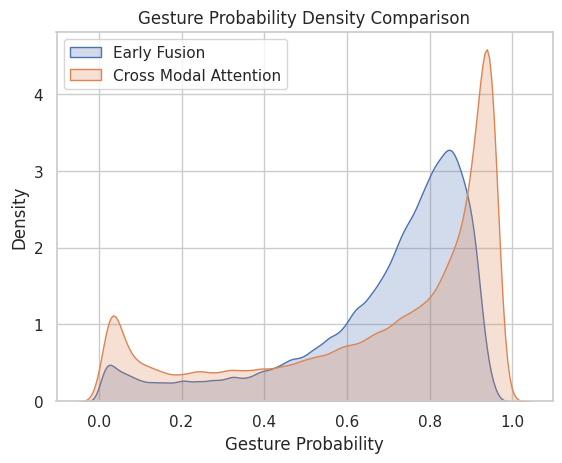}
        \caption{Distributions of gesture predictions in early and cross-modal fusion approaches.}
        \label{fig:probability_dists}
    \end{minipage}%
    \hfill
    \begin{minipage}{0.65\linewidth}
        \centering
        \includegraphics[width=\linewidth]{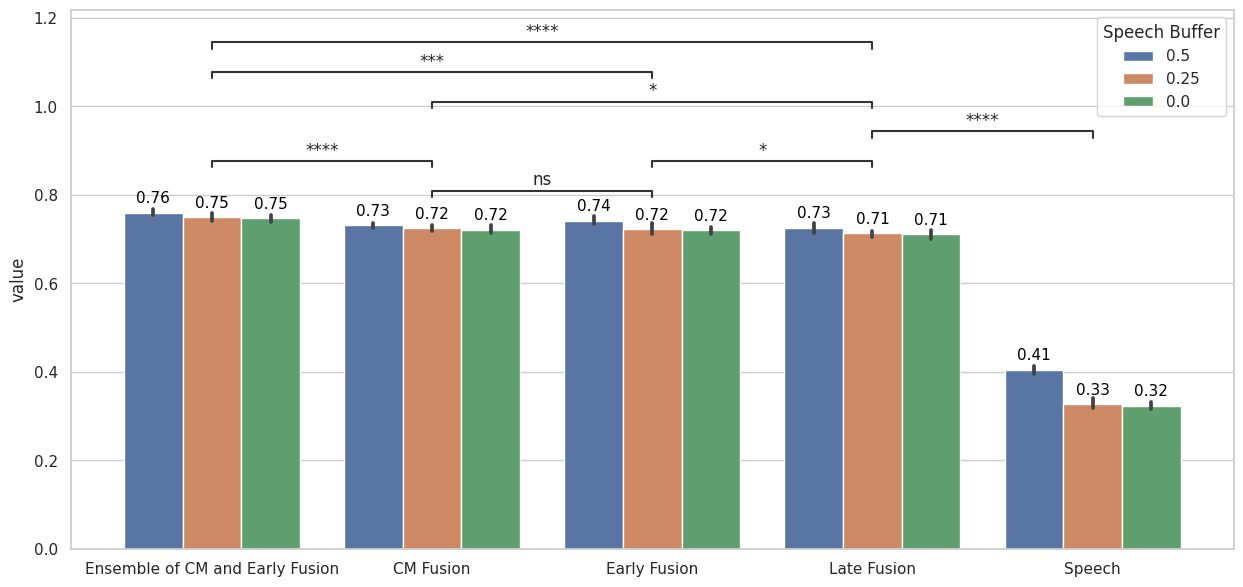}
        \caption{MAP of fusion and speech models with different speech buffers. Note that ensembling early and cross-modal models provides the best MAP, significantly outperforming other models.}
        \label{fig:ensambling_of_cross_early_fusion}
    \end{minipage}
\end{figure}

\textit{Second,} compared to late fusion, in which modalities are ensembled at the decision level, cross-modal and early fusion approaches show better performance. 
For instance, the cross-modal fusion approach without speech buffer has higher F1 and MAP (68.7 and 72.2) than late fusion (67.1 and 71.1). 
As shown in Figure \ref{fig:ensambling_of_cross_early_fusion}, the cross-modal and early fusion approaches' MAP is significantly better than the MAP late fusion model. Interestingly, the late fusion approach can achieve scores similar to those of cross-modal fusion only when a high speech buffer of 500ms is used. This suggests that integrating and \textit{aligning} speech and vision signals is more effective for accurate gesture detection than simply ensembling their outcomes.

Third, across all fusion techniques, an increase in the speech buffer size generally leads to better performance metrics (see Figure \ref{fig:ensambling_of_cross_early_fusion} and Table \ref{tab:all_vggish_results}). This implies that a larger temporal window for speech data contributes to more effective alignment and integration with gestures, enhancing detection metrics. However, the significance of the speech buffer is more noticeable in the late fusion technique as it offers a heuristic for aligning speech and visual signals, which is comparatively less necessary in the cross-modal and early fusion approaches. This is because the latter methods have their own mechanisms for aligning and integrating the two cues over time.

In summary, our findings highlight the advantage of multimodal approaches, showing that speech has complementary information for co-speech gesture detection. We observe that combining information from speech and visual cues, especially through early and cross-modal fusion approaches, is more powerful than using unimodal approaches. While all methods benefit from an expanded speech buffer, the impact is more pronounced in late fusion. In particular, early and cross-modal fusion approaches exhibit comparable performance (with insignificant differences as shown in Figure \ref{fig:ensambling_of_cross_early_fusion}), while both outperform late fusion significantly.

\section{Further Analyses}
\label{sect:analysis} 
This section analyses the results of the bimodal models and checks how various fusion approaches, even if they give similar results, have differences that can be further exploited. We also study how models' predictions correlate with low-level speech features and show a case of how speech models use such features using a visualisation technique.

\subsection{Fusion Approaches}
\subsubsection{Comparative Analysis and Ensembling of Cross-Modal and Early Fusion}
Although cross-modal and early fusion approaches yield similar results with insignificant differences in terms of F1 and MAP, their mechanisms and training trajectories to integrate speech and visual signals related to co-speech gestures differ. 
We hypothesise that they capture different information when combining speech and gestures. For example, we notice that the distribution of their predictions (shown in Figure \ref{fig:probability_dists}) and the predicted samples differ significantly.
For the positive class samples (i.e., gesture), the median confidences in gesture prediction in cross-modal and early fusion are 0.79 and 0.75, respectively. The distributions in the two models differ significantly (Mann–Whitney U = 2444082115.5, n1 = n2 = 66049, $P  \ll0.00001$ two-tailed). In addition, the two approaches differ in $16\%$ of their predictions, and both make mistakes in around $29\%$ of the gesture samples. These errors, however, do not necessarily occur in the same samples. For this reason, we check to what extent ensembling these two fusion approaches helps the performance. We ensemble the cross-modal and early fusion predictions by taking their averages. 
This procedure improves detection performance to 76.0 MAP and 71.3 F1 when using a speech buffer of 500ms. These results are much better than those obtained when early and cross-modal fusion are taken independently (refer to Figure \ref{fig:ensambling_of_cross_early_fusion}).

\begin{figure}
    \centering
    \includegraphics[width=1\linewidth]{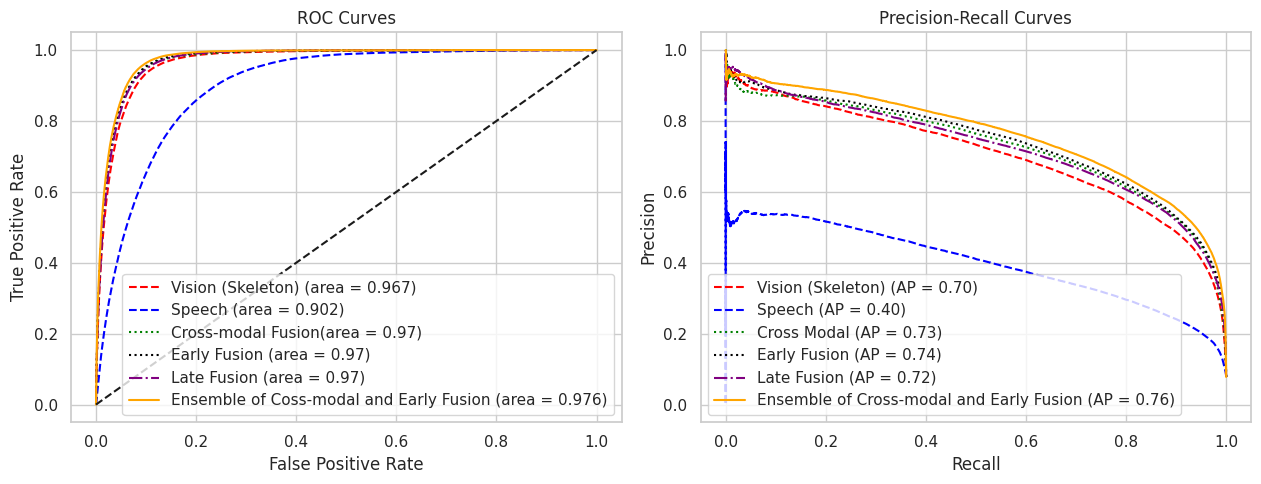}
    \caption{ROC and PR curves for the fusion approaches and the unimodal baseline models.}
    \label{fig:pr_roc_curves}
\end{figure}

\subsubsection{Bimodal Detection Improves Precision}
The fusion approaches notably improve the precision of gesture detection by significantly reducing the false positive rate observed in unimodal models. In particular, bimodal detection exhibits enhanced precision compared to vision-only methods, achieving a better balance between precision and recall. 
We observe that including speech information plays a crucial role in this enhancement by discriminating between actual gestures and mere ``movements'' not accompanied by speech.
Furthermore, the precision-recall and ROC curves are shown in Figure \ref{fig:pr_roc_curves}, illustrating the advantage of bimodal detection over the unimodal baselines.
The figure shows that ensembling early and cross-modal fusion methods is the most effective fusion strategy.

\begin{figure}
    \centering
    \includegraphics[width=0.99\linewidth]{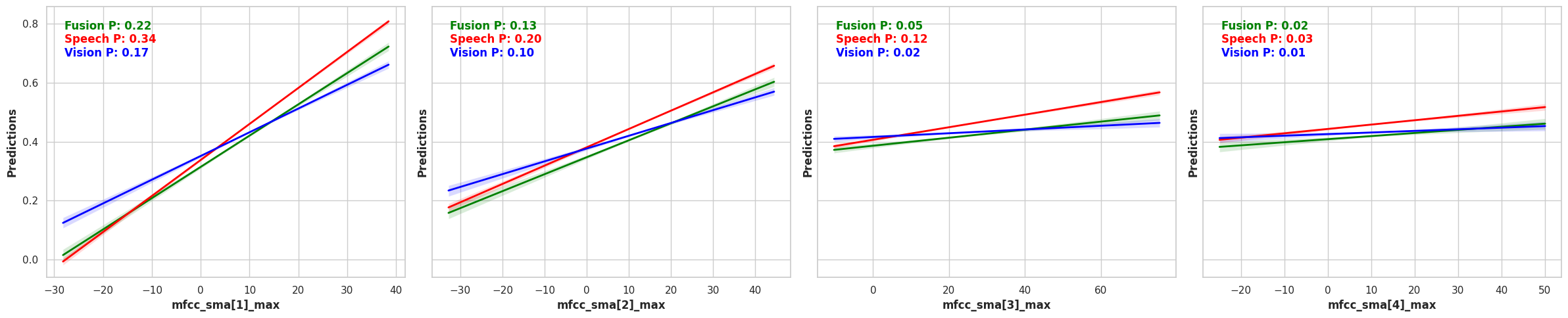}
    \caption{Correlation of maximum of MFCCs with speech, vision, and cross-modal confidence in predicting gestures.}
    \label{fig:mfcc_1_max_correlation}
\end{figure}

\subsection{Speech Low-level Features and Models' Confidence in Gesture Prediction}\label{sect:confidence_analysis}
In section \ref{sect:speech_low_level_features}, we demonstrated that some of the frequency-related speech features, such as MFCCs (particularly the first and second coefficients) and log of $F0$ with respect to its harmonics, are clearly higher when a gesture accompanies speech. Previous studies have proven that hand-crafted MFCCs and $F0$ features can effectively be used to generate \cite{hasegawa2018evaluation, kucherenko2019analyzing} and predict \cite{yunus2021sequence} gestures from speech alone. In this analysis, we aim to investigate to what extent the confidence in gesture prediction by the speech and fusion models is associated with these low-level features.

We observe that when speech models are used, confidence in gesture prediction is positively correlated with speech features that are also associated with speech accompanied by gestures. Higher MFCC values, especially for MFCC[1]\&[2], correlate with increased confidence in gesture prediction. 
The speech model has the highest Spearman correlation coefficient ($\rho=0.34$) for Maximum of MFCC[1], followed by the fusion (cross-modal) model ($\rho=0.22$, $p\ll0.00001$) and vision model ($\rho=0.17$, $p\ll0.00001$).
These results are illustrated in Figure \ref{fig:mfcc_1_max_correlation}.
According to the graph, the correlation values for the maximum value of the second MFCC decrease for all three models. However, there is a weak positive correlation ($\rho=0.1$, $p\ll0.00001$) with only the speech model for the third MFCC.
Nonetheless, the correlation values are insignificant for the fourth MFFC.
Additionally, the analysis includes the logarithm of the ratio of the $F0$ relative to the first and third harmonics. Here, the speech model again shows a higher positive Spearman correlation ($\rho=0.26$, $p\ll0.00001$) with gesture predictions, followed by the fusion model ($\rho=0.17$, $p\ll0.00001$) and the vision one ($\rho=0.13$, $p\ll0.00001$).
The results show a clear pattern, consistent with the outcome of the analysis in Section \ref{sect:speech_low_level_features}: speech-based models have higher confidence in gesture prediction when higher MFCC values and the log ratio of F0 to its harmonics are present.

\begin{figure}[ht]
    \centering
    \begin{minipage}{0.47\linewidth}
        \centering
        \includegraphics[width=\linewidth]{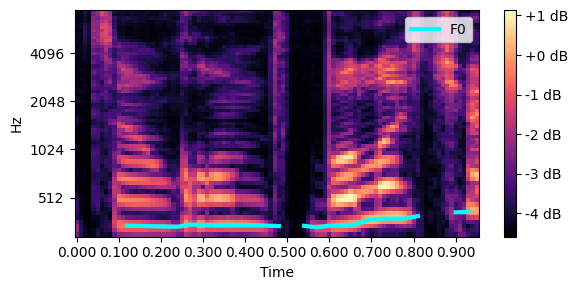}
        \caption{Mel-spectrogram where the $F0$ contours are marked across time.}
        \label{fig:spectrogram}
    \end{minipage}
    \hfill
    \hfill
    \begin{minipage}{0.44\linewidth}
        \centering
        \includegraphics[width=\linewidth]{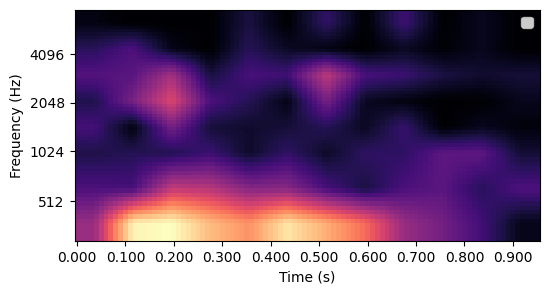}
        \caption{The spectrogram's class activation map shows that the Mel band and times corresponding to segments where $F0$ contours are activated.}
        \label{fig:cam}
    \end{minipage}%

\end{figure}

\subsection{Exploratory Analysis with Class Activation Maps}
In this analysis, we visualise the low-level features that the speech backbone model pays attention to when used for gesture detection, focusing on the fundamental frequency $F0$. 
The speech model operates on the Mel-spectrogram, which is a frequency-based speech representation. 
Figure \ref{fig:spectrogram} illustrates the $F0$ contours in a Mel-spectrogram, which we extract through pYIN algorithm \cite{mauch2014pyin} in Librosa \cite{mcfee2015librosa}. 
Utilizing Class Activation Maps (CAMs), we check whether the CNN model activations correspond to regions in a Mel-Spectrogram that are important for gesture detection. 
In this technique, the score of the predicted class (i.e., gesture or neutral) is projected back to the convolutional layers to generate the CAMs. We specifically employ Grad-CAM proposed by \cite{selvaraju2017grad}, which utilizes gradient weights to produce a localization map of the significant regions to predict the class of interest from the input spectrogram.

We find that the last layer of the speech backbone model (i.e., VGGish's sixth convolutional layer) carries the most activations corresponding to $F0$. The activation is illustrated in Figure \ref{fig:cam}. 
Although these localization maps provide a window into which features are being exploited by a neural network, we do not find significant differences between the activations of the $F0$ regions for gestural and non-gestural time windows. We leave this interesting direction for future work.
\section{Discussion}
\label{sect:discussion}
This study shows that detecting co-speech gestures can be improved with models that exploit both speech and vision. In this section, we discuss some of the main take-aways of our work and highlight some open questions. 


\paragraph{The predictive power of speech features}
Our study has revealed significant differences in low-level speech frequency features when gestures accompany speech. It has also shown that a deep learning model with only speech information can predict the presence of co-speech gestures above the chance level. Furthermore, adding speech input to a model that includes visual skeletal information significantly improves performance over a model that exploits only visual input. Our results, therefore, demonstrate that speech has predictive power when it comes to detecting gestures. Nonetheless, it remains unclear how precisely different speech features contribute to this predictive power. Our analyses have shed some light on this issue: For example, we have seen that when gestures accompany speech, certain features in the acoustic frequency tend to be higher (e.g., MFCC[1]), which positively correlates with the models' confidence in predicting the gesture. Similarly, the class activation maps suggest that the model may pay attention to $F0$, which supports the literature on multimodal prosody \cite{wagner2014gesture, pouw2020energy}. However, using only low-level speech features limits the inferences we can draw about what exactly in speech is aiding gesture detection. For instance, it is likely that semantic and syntactic information related to the lexical affiliate (e.g., the type of part of speech being uttered) plays a role and that the predictive power of low-level features is driven by their indirect relation with such higher-order information. At this point, we can only conclude that a more detailed investigation is needed to understand how different properties of speech relate to the presence of gestures.

\paragraph{Temporal alignment between speech and visual information.} Studies have shown that speech and gesture onsets are temporally coordinated, though not always perfectly aligned. Our experimentation with heuristic buffers and advanced Transformer encoder models highlights the importance of temporally aligning the two signals in an asymmetric way, such that speech context after the gesture is taken into account to predict gesture occurrence. This finding will inform gesture analysis studies to carefully address the asynchrony, e.g., in gesture generation. However, our analysis did not investigate whether previous rather than post-gesture speech segments can influence gesture detection as well. Future research in this direction could provide valuable insights into the predictive relationship between speech and subsequent gestural expressions.

\paragraph{Single and multi-stream learning and fusion.} In line with the literature on machine learning for vision and speech tasks, our findings show the advantage of integrating and aligning modalities over simple ensemble methods. Similar to the results from multimodal emotion recognition studies \cite{rajan2022cross}, we find that both single-stream (self-attention based) and multi-stream (cross-attention based) models yield comparable overall performance outcomes. However, an interesting aspect is that these models tend to differ in their confidence predictions and predict different sample sets. A similar phenomenon has been noted in vision and language tasks, where diverse processing streams contribute uniquely to the task's overall performance, as detailed in studies like the one conducted by Baltrušaitis \etal \cite{baltruvsaitis2018multimodal}.

\section{Conclusion}
\label{sect:conclusion}
Gestures are an inherent component of human communication, particularly in face-to-face interaction, where they tend to co-occur with speech. Current automatic gesture detection approaches, however, concentrate on a finite set of silent gestures, ignoring speech cues. This paper addresses this gap by investigating co-speech hand gestures in naturalistic conversation and proposing methods to exploit synchronies between skeletal and speech information for the gesture detection task. 
Through novel approaches, we successfully tackle several challenges, including the great variability in form and duration of co-speech gestures and the misalignment of speech and gestural cues. We also handle technical challenges as both data streams have different sampling rates, and their latent space does not correlate. 
Our results indicate that integrating both speech and visual cues enhances detection performance. Our findings also highlight the importance of not only combining speech and gestures but also ensuring their temporal alignment by means of a speech buffer or by using Transformer encoder models within a multimodal fusion approach. In addition, we carry out several analyses that suggest that the speech-based models exploit low-level features such as the mean and maximum of MFCC[1] and MFCC[2] and the fundamental frequency $F0$ to make gesture predictions. In conclusion, we believe that this study advances our understanding of co-speech gestures together with our capability to automatically detect them in multimodal communication.

\begin{acks}
This work was funded by the Dutch Research Council (NWO) through a Gravitation grant (024.001.006) to the Language in Interaction consortium.
Raquel Fern\'andez is supported by the European Research Council (ERC CoG grant agreement 819455).
Further funding was provided by the DFG (project number 468466485) and the Arts and Humanities Research Council (grant reference AH/W010720/1) to Peter Uhrig and Anna Wilson (University of Oxford).
We thank the Dialogue Modelling Group members at UvA, especially Alberto Testoni and Sandro Pezzelle, for their valuable feedback. 
The authors gratefully acknowledge the scientific support and HPC resources provided by the Erlangen National High-Performance Computing Center (NHR@FAU) of the Friedrich-Alexander-Universität Erlangen-Nürnberg (FAU) under the NHR project b105dc to Peter Uhrig. NHR funding is provided by federal and Bavarian state authorities. NHR@FAU hardware is partially funded by the German Research Foundation (DFG) – 440719683.
\end{acks}

\bibliographystyle{ACM-Reference-Format}
\bibliography{main}

\appendix
\pagebreak
\section{Speech Features}
\subsection{Distribution of Speech Features}\label{app:speech_feat}
\begin{table}[h]
\scalebox{0.8}{
\begin{tabular}{l|cr|cr|cr|cr}
\toprule
        feature name &  \multicolumn{2}{c}{Is speech feature different?}  &  \multicolumn{2}{c}{Speech model corr.} &  \multicolumn{2}{c}{Cross-model corr.} &  \multicolumn{2}{c}{Vision model corr.} \\
        
         &  t & $p-value$  &  $\rho$ &  $p-value$  &  $\rho$ &  $p-value$  &  $\rho$ &  $p-value$ \\
\midrule
     \rowcolor{green!10} MFCC[1] max &    31.14087 &  $\ll0.00001$ &        0.346 &   $\ll0.00001$ &             0.216 &        $\ll0.00001$ &        0.168 &   $\ll0.00001$ \\
    \rowcolor{green!10} logRelF0-H1-A33 amean &    20.96735 &  $\ll0.00001$ &        0.262 &   $\ll0.00001$ &             0.169 &        $\ll0.00001$ &        0.133 &   $\ll0.00001$ \\
     \rowcolor{green!10} MFCC[2] max &    18.14325 &  $\ll0.00001$ &        0.202 &   $\ll0.00001$ &             0.131 &        $\ll0.00001$ &        0.101 &   $\ll0.00001$ \\
   \rowcolor{green!10} MFCC[1] amean &    15.77846 &  $\ll0.00001$ &        0.209 &   $\ll0.00001$ &             0.120 &        $\ll0.00001$ &        0.091 &   $\ll0.00001$ \\
   \rowcolor{green!10} MFCC[2] amean &    14.44394 &  $\ll0.00001$ &        0.092 &   $\ll0.00001$ &             0.062 &        $\ll0.00001$ &        0.041 &   $\ll0.00001$ \\
      \rowcolor{green!10} F0env amean &     9.32535 &  $\ll0.00001$ &        0.090 &   $\ll0.00001$ &             0.070 &        $\ll0.00001$ &        0.057 &   $\ll0.00001$ \\
       \rowcolor{green!10} F0env min &     8.02577 &  $\ll0.00001$ &        0.086 &   $\ll0.00001$ &             0.061 &        $\ll0.00001$ &        0.051 &   $\ll0.00001$ \\
  \rowcolor{green!10} F0env kurtosis &     6.51304 &  $\ll0.00001$ &        0.146 &   $\ll0.00001$ &             0.094 &        $\ll0.00001$ &        0.080 &   $\ll0.00001$ \\
     \rowcolor{green!10} MFCC[2] min &     5.41853 &  $\ll0.00001$ &        0.002 &   0.70353 &             0.019 &        0.00004 &        0.019 &   0.00005 \\
     \rowcolor{green!10} MFCC[4] max &     5.13618 &  $\ll0.00001$ &        0.033 &   $\ll0.00001$ &             0.022 &        $\ll0.00001$ &        0.016 &   0.00073 \\
\rowcolor{green!10} MFCC[2] skewness &     4.41855 &  0.00001 &        0.032 &   $\ll0.00001$ &             0.033 &        $\ll0.00001$ &        0.038 &   $\ll0.00001$ \\
     \rowcolor{green!10} MFCC[3] max &     3.42256 &  0.00062 &        0.117 &   $\ll0.00001$ &             0.044 &        $\ll0.00001$ &        0.023 &   $\ll0.00001$ \\
        \rowcolor{red!10}F0 amean &     0.73532 &  0.46215 &        0.096 &   $\ll0.00001$ &             0.047 &        $\ll0.00001$ &        0.033 &   $\ll0.00001$ \\
\rowcolor{red!10}MFCC[4] kurtosis &     0.71139 &  0.47685 &       -0.031 &   $\ll0.00001$ &            -0.002 &        0.71263 &        0.004 &   0.43969 \\
      \rowcolor{red!10} F0env max &    -0.66383 &  0.50680 &       -0.029 &   $\ll0.00001$ &            -0.003 &        0.53054 &       -0.003 &   0.48146 \\
          \rowcolor{blue!10}F0 max &    -2.21193 &  0.02698 &       -0.047 &   $\ll0.00001$ &            -0.016 &        0.00062 &       -0.015 &   0.00120 \\
\rowcolor{blue!10} MFCC[3] kurtosis &    -2.22375 &  0.02617 &       -0.072 &   $\ll0.00001$ &            -0.034 &        $\ll0.00001$ &       -0.024 &   $\ll0.00001$ \\
\rowcolor{blue!10}MFCC[2] kurtosis &    -3.07645 &  0.00210 &       -0.019 &   0.00005 &            -0.011 &        0.02448 &       -0.011 &   0.01911 \\
          \rowcolor{blue!10}F0 min &    -3.38433 &  0.00071 &       -0.028 &   $\ll0.00001$ &            -0.024 &        $\ll0.00001$ &       -0.020 &   0.00002 \\
\rowcolor{blue!10}MFCC[1] skewness &    -3.76709 &  0.00017 &       -0.123 &   $\ll0.00001$ &            -0.052 &        $\ll0.00001$ &       -0.034 &   $\ll0.00001$ \\
     \rowcolor{blue!10}F0 skewness &    -4.22045 &  0.00002 &       -0.129 &   $\ll0.00001$ &            -0.057 &        $\ll0.00001$ &       -0.043 &   $\ll0.00001$ \\
\rowcolor{blue!10}MFCC[3] skewness &    -4.24599 &  0.00002 &       -0.064 &   $\ll0.00001$ &            -0.041 &        $\ll0.00001$ &       -0.034 &   $\ll0.00001$ \\
   \rowcolor{blue!10}MFCC[3] amean &    -5.12520 &  $\ll0.00001$ &        0.034 &   $\ll0.00001$ &            -0.004 &        0.42029 &       -0.012 &   0.01132 \\
     \rowcolor{blue!10}F0 kurtosis &    -6.32490 &  $\ll0.00001$ &       -0.144 &   $\ll0.00001$ &            -0.071 &        $\ll0.00001$ &       -0.060 &   $\ll0.00001$ \\
\rowcolor{blue!10}MFCC[1] kurtosis &    -6.44900 &  $\ll0.00001$ &       -0.109 &   $\ll0.00001$ &            -0.069 &        $\ll0.00001$ &       -0.056 &   $\ll0.00001$ \\
   \rowcolor{blue!10}MFCC[4] amean &    -7.43256 &  $\ll0.00001$ &       -0.162 &   $\ll0.00001$ &            -0.085 &        $\ll0.00001$ &       -0.063 &   $\ll0.00001$ \\
     \rowcolor{blue!10}MFCC[1] min &    -9.53406 &  $\ll0.00001$ &       -0.137 &   $\ll0.00001$ &            -0.075 &        $\ll0.00001$ &       -0.053 &   $\ll0.00001$ \\
\rowcolor{blue!10}MFCC[4] skewness &   -10.43039 &  $\ll0.00001$ &       -0.056 &   $\ll0.00001$ &            -0.048 &        $\ll0.00001$ &       -0.037 &   $\ll0.00001$ \\
  \rowcolor{blue!10}F0env skewness &   -10.87089 &  $\ll0.00001$ &       -0.160 &   $\ll0.00001$ &            -0.095 &        $\ll0.00001$ &       -0.081 &   $\ll0.00001$ \\
     \rowcolor{blue!10}MFCC[3] min &   -15.37323 &  $\ll0.00001$ &       -0.159 &   $\ll0.00001$ &            -0.099 &        $\ll0.00001$ &       -0.081 &   $\ll0.00001$ \\
     \rowcolor{blue!10}MFCC[4] min &   -18.57138 &  $\ll0.00001$ &       -0.233 &   $\ll0.00001$ &            -0.139 &        $\ll0.00001$ &       -0.106 &   $\ll0.00001$ \\
\bottomrule
\end{tabular}
}
\caption{Hand-crafted features extracted through Speech Processing tool: OpenSmile \cite{eyben2015opensmile}. In this table, features highlighted in green tend to have significantly higher values when gestures accompany speech, while features in blue tend to have significantly lower values when gestures accompany speech. Features highlighted in red are indifferent.}
\end{table}

\end{document}